\documentclass[journal]{IEEEtran}
\ifCLASSINFOpdf
\else
\fi
\usepackage{cite}
\usepackage{amsmath,amssymb,amsfonts}
\usepackage{graphicx}
\usepackage{array}
\usepackage{mdwmath}
\usepackage{mdwtab}
\usepackage{eqparbox}
\usepackage{url}
\usepackage{tabularx}
\usepackage{mathrsfs}
\usepackage{rsfso}
\usepackage{multirow}
\usepackage{amssymb}
\usepackage{multirow, booktabs}
\usepackage{fancyhdr}
\usepackage{amsmath}
\usepackage{algorithm}
\usepackage{algpseudocode}

\begin{document}
%
% paper title
% Titles are generally capitalized except for words such as a, an, and, as,
% at, but, by, for, in, nor, of, on, or, the, to and up, which are usually
% not capitalized unless they are the first or last word of the title.
% Linebreaks \\ can be used within to get better formatting as desired.
% Do not put math or special symbols in the title.
\title{Semi-Supervised Deep Domain Adaptation for Predicting Solar Power  Across Different Locations }

\author{

% \IEEEauthorblockN{Md Shazid Islam\IEEEauthorrefmark{1}, A S M Jahid Hasan \IEEEauthorrefmark{2}, Md Saydur Rahman\IEEEauthorrefmark{1}, Jubair Yusuf\IEEEauthorrefmark{4},\\ Md Saiful Islam Sajol\IEEEauthorrefmark{3},and  Md Taukir Azam Chowdhury \IEEEauthorrefmark{1}}

% \IEEEauthorblockA{\IEEEauthorrefmark{1}University of California Riverside,California, USA,\IEEEauthorrefmark{2}North South University, Dhaka, Bangladesh,\\\IEEEauthorrefmark{4}Sandia National Laboratories, Albuquerque, New Mexico, USA, \IEEEauthorrefmark{3}Louisiana State University, Louisiana, USA\\Email: misla048@ucr.edu, jahid.hasan12@northsouth.edu, mrahm054@ucr.edu,\\jyusuf177@gmail.com, msajol1@lsu.edu, mchow068@ucr.edu}

\IEEEauthorblockN{
Md Shazid Islam\IEEEauthorrefmark{1}, 
A S M Jahid Hasan\IEEEauthorrefmark{2}, 
Md Saydur Rahman\IEEEauthorrefmark{1}, 
Md Saiful Islam Sajol\IEEEauthorrefmark{3}
}

\IEEEauthorblockA{
\IEEEauthorrefmark{1}University of California, Riverside, California, USA\\
\IEEEauthorrefmark{2}North South University, Dhaka, Bangladesh\\
\IEEEauthorrefmark{3}Louisiana State University, Louisiana, USA\\
Email: misla048@ucr.edu, jahid.hasan12@northsouth.edu, mrahm054@ucr.edu, msajol1@lsu.edu
}
}

\maketitle

% As a general rule, do not put math, special symbols or citations
% in the abstract or keywords.
\begin{abstract}
% Accurate solar power generation forecasting is crucial for integrating renewable energy sources across diverse geographic regions. However, solar power prediction models often face domain shift challenges due to varying weather conditions across locations, limiting their transferability. This paper presents a semi-supervised deep domain adaptation framework to address these issues, enabling accurate predictions with minimal labeled data from the target domain. Our approach involves training a deep convolutional neural network on a source location's data and adapting it to the target location using a source-free, teacher-student model configuration. The teacher-student model leverages consistency and cross-entropy loss for semi-supervised learning, ensuring effective adaptation without requiring access to source data during deployment. Evaluations on datasets from California, Florida, and New York demonstrate that our method significantly improves prediction accuracy, with up to 11.36 $\%$ gains over non-adaptive models with annotation on only 20 $\%$ data in the target domain. Additionally, our model reduces annotation costs while maintaining robust performance under various supervision levels, showing promise for scalable, location-agnostic solar power forecasting.

Accurate solar generation prediction is essential for proper estimation of renewable energy resources across diverse geographic locations. However, geographical and weather features vary from location to location which introduces domain shift - a major bottleneck to develop location-agnostic prediction model. As a result, a machine-learning model which can perform well to predict solar power in one location, may exhibit subpar performance in another location. Moreover, the lack of properly labeled data and storage issues make the task even more challenging. In order to address domain shift due to varying weather conditions across different meteorological regions, this paper presents a semi-supervised deep domain adaptation framework, allowing accurate predictions with minimal labeled data from the target location. Our approach involves training a deep convolutional neural network on a source location's data and adapting it to the target location using a source-free, teacher-student model configuration. The teacher-student model leverages consistency and cross-entropy loss for semi-supervised learning, ensuring effective adaptation without any source data requirement for prediction. With annotation of only $20 \%$ data in the target domain, our approach exhibits an improvement upto $11.36 \%$, $6.65 \%$, $4.92\%$ for California, Florida and New York as target domain, respectively in terms of accuracy in predictions with respect to non-adaptive approach.

% The results exhibit that proposed approach can significantly improve prediction accuracy, achieving up to an 11.36\% increase over non-adaptive models with only 20 $\%$ annotated data in the target domain, evaluated on datasets from California, Florida, and New York.

\end{abstract}

% Note that keywords are not normally used for peerreview papers.

% For peer review papers, you can put extra information on the cover
% page as needed:
% \ifCLASSOPTIONpeerreview
% \begin{center} \bfseries EDICS Category: 3-BBND \end{center}
% \fi
%
% For peerreview papers, this IEEEtran command inserts a page break and
% creates the second title. It will be ignored for other modes.
\IEEEpeerreviewmaketitle


\section{Introduction}

% \textcolor{red}{1. Why Solar Power Prediction is important?} \\
% \textcolor{red}{2. What is the impact of Location in Solar power generation?}\\
% \textcolor{red}{3. What is Domain Shift?}\\
% \textcolor{red}{4. How Domain Shift occurs in Solar power prediction? (explain with a figure)}\\
% \textcolor{red}{5. Superrvised, Semi supervised and Unsupervised Adaptation}\\
% \textcolor{red}{6. Our contribution }\\
%\textcolor{red}{Check/ Rewrite all captions}

\begin{figure*}
\centering
\includegraphics[width=\textwidth]{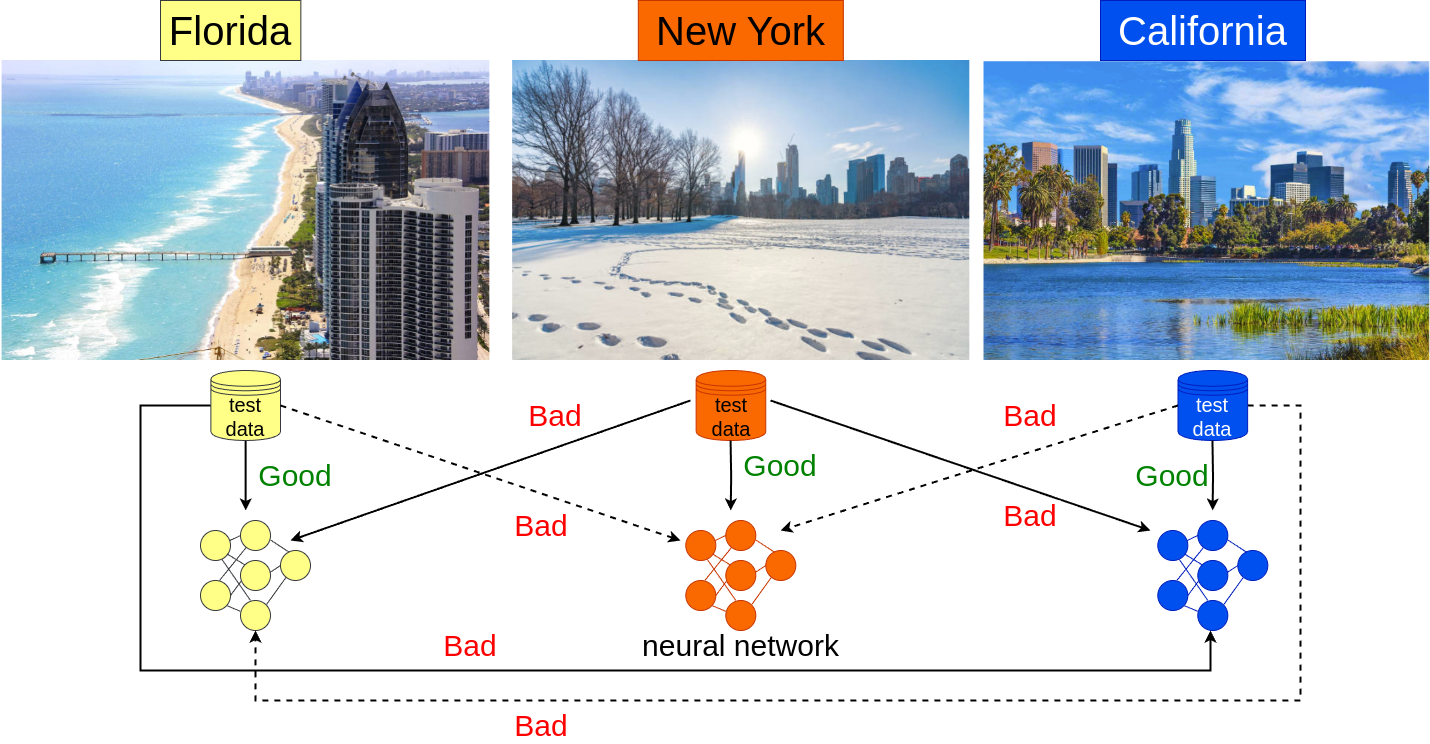}
\caption{Domain shift issue: Florida (FL), New York (NY) and California (CA) have different weather features.  Three different colours have been used to demonstrate three different locations (FL-Yellow, NY-Orange, CA-Blue). Three different neural networks are trained on the data of distinct locations which are shown by the same color of that area. After training, we observe that model trained on data of FL, shows good performance on the test data of FL. However, it exhibits bad performance when it is tested against the test data of different locations : NY and CA. Same incident occurs for both NY and CA.}
\label{domain_shift}
\end{figure*}

% The fast population growth along with advancement in the world economy has brought about an increase in energy usage everywhere. The US Energy Administration (EIA) estimates that by 2050 there will be more than a 50\% increment in energy usage worldwide \cite{b1}. However, relying on fossil fuels fully to meet this increased demand will cause more Greenhouse Gas (GHG) emissions as well as faster depletion of useful resources. Renewable energy sources offer the best solution to this problem as they are abundant and clean. So, renewable energy is being integrated into the grid more in both developed and developing countries alike. According to the International Energy Agency (IEA), by 2025 the combined energy generation from solar and wind will surpass the generation from coal \cite{b2}. \\

Rapid population growth, combined with advancements in the global economy, has led to a significant rise in energy consumption worldwide. According to a projection by the U.S. Energy Information Administration (EIA), worldwide energy consumption may rise by as much as 50\% by 2050. \cite{b1}. However, since fossil fuel resources are limited and will eventually be depleted, substitute energy sources are needed to meet this growing demand. Renewable energy sources present an alternative solution to this energy demand, with increased use of natural resources when delivering energy. As a result, renewable energy is increasingly being adopted into the grid as a power generation source in both developed and developing countries. International Energy Agency (IEA) reports that, the combined energy generation from solar and wind will surpass the generation from coal in 2025 \cite{b2}. Especially, solar photovoltaic (PV) energy stands out in the renewable energy sector, with module prices dropping by 93\% over the past decade and the levelized cost of energy for utility-scale PV decreasing by 85\% in the same period \cite{irenaSolarEnergy}. This surge in solar power is further supported by the fact that, in 2023, solar PV alone contributed to three-quarters of the total renewable capacity addition \cite{ieaSolar}. 
% Solar PV has the potential to play a pivotal role in achieving the net-zero carbon emission goal set by the United Nations (UN) by 2050 \cite{unZeroCoalition}.

% Unfortunately, the availability and reliability of weather data are major concerns in utilizing weather variables for any applications all over the world. In particular, the credibility of weather data in developing countries are not as sufficient as the data in developed countries to train and test the machine learning models. 

% As mentioned earlier, developing countries are also inclining towards higher renewable energy integration into the grid. Thus, the task of solar generation estimation is imperative in these countries.

However, the problem with many renewable energy resources is that they are highly dependent on the weather variables and intermittent in nature. When weather characteristic variables and their values are known, renewable generation (e.g., from solar PV and wind) can be estimated using statistical or machine learning approaches. Additionally, weather variables exhibit strong location dependence and can vary substantially based on the meteorological region of interest.
 The statistical distribution of weather data varies across different locations, a phenomenon referred to in data science as \emph{domain shift} \cite{stacke2020measuring}. Due to domain shift across different locations, machine learning models are not universally applicable, and a single model cannot be used for any arbitrary location. Consequently, each location of interest requires the model to be trained individually using its specific weather dataset. Otherwise, using a model trained on one dataset of one location to estimate the solar generation of some other location will introduce erroneous results due to high bias towards the source dataset. Therefore, domain shift can complicate solar generation prediction when analyzing multiple locations with distinct meteorological profiles. Fig. \ref{domain_shift} demonstrates this phenomenon using three U.S. states, each representing different weather regimes. California (CA) exhibits Mediterranean weather patterns, Florida (FL) displays humid subtropical conditions, and New York (NY) shows humid continental characteristics, highlighting the significant variations in atmospheric conditions across these regions.

Developing separate models for each location can address this issue, but it would require building every model from scratch, leading to higher computational and time cost. Domain adaptation, a special form of transfer-learning,  provides an effective approach to overcome data limitations by transferring knowledge between related domains \cite{b27,b28,b29, Tipu, PINTO, sumon}. In this framework, models trained on source domain data (e.g., weather patterns from one region) can be adapted to perform accurate solar generation predictions in a target domain (another meteorological region), despite distributional differences causing domain shift between the datasets.

Domain adaptation is generally categorized into two types: with-source adaptation \cite{hoffman2018algorithms} and source-free adaptation \cite{b28}. In with-source adaptation, the training data used to initially train the model is required during the adaptation process, meaning both source data and labels must be accessible. However, this approach demands additional memory to store and utilize the source data during adaptation. In contrast, source-free adaptation eliminates the need for source data during the adaptation step, thereby bypassing the requirement for extra memory storage. In this work, we incorporate a source-free domain adaptive setting to predict solar power generation of different regions that have wide variations in their weather profiles. This prediction methodology becomes location-agnostic while being computationally efficient, meaning a reduction in time and memory storage.

% \textcolor{red}{Motivate the source-free setting}

% Data collection, refinement, storage, labeling pose huge challenges in machine learning research. Recent research shows that the majority of engineering efforts in AI and machine learning projects are dedicated to data preparation and labeling. \cite{sambasivan2021everyone}. Given that data is mostly annotated manually, this process is both labor-intensive and prone to errors. As mentioned earlier, in many developing countries, there is a scarcity of reliable data, making it even more challenging to obtain properly labeled datasets. Leveraging unsupervised or semi-supervised domain adaptation can help resolve this issue. Unsupervised Domain Adaptation (UDA) performs domain adaptation without any labeled data in the target dataset, while Semi-supervised Domain Adaptation uses only a small amount of labeled data in the target dataset \cite{liu2022,yu2023}. These methods are forms of unsupervised or semi-supervised learning designed to handle domain shifts between source and target datasets, where the target data may be partially labeled or entirely unlabeled. Additionally, training a machine learning model requires huge amount of data. In real world storing this huge amount of data requires hardwire support which is expensive. Hence we adopt the source-free method for domain adaptation which bypasses the requirement of source-data mitigating the requirement of additional storage.

One of the major challenges in machine learning is data labeling. Recent research shows that the majority of engineering efforts in AI and machine learning projects are dedicated to data preparation and labeling. \cite{sambasivan2021everyone}. Given that data is mostly annotated manually, this process is both labor-intensive and prone to errors. As mentioned earlier, in many developing countries, it is challenging to obtain properly labeled datasets. Leveraging unsupervised or semi-supervised domain adaptation can help resolve this issue. Unsupervised Domain Adaptation (UDA) performs domain adaptation without any labeled data in the target dataset, while Semi-supervised Domain Adaptation uses only a small amount of labeled data in the target dataset \cite{liu2022,yu2023}. Motivated by these observations, we consider a source-free domain adaptation method with limited supervision in the target side (semi-supervised). A model is first trained on a particular location with respective location-based dataset (source domain). Then the trained model is adapted to the weather features of a separate location (target domain) where only a small portion of data is labeled. The semi-supervised source-free adaptation reduces both annotation and storage costs, respectively. The contributions of this paper are as follows: 

\begin{itemize}
    \item Incorporating a teacher-student model-based training scheme in source-free domain adaptation approach to predict solar power (kW) generation from weather features, effectively addressing domain shifts caused by varying weather conditions across different locations.

    \item Alleviating the manual annotation burden by adopting a semi-supervised domain adaptation setting, which significantly reduces annotation costs with small performance degradation as a trade-off.

    % \item Reducing computational cost and running time by effective feature selection and adaptation for the large dataset prediction task.

    \item Making the proposed method storage efficient by utilizing source-free constraint which means source data is not available during adaptation.

    \item With annotation of only $20 
    \%$ data in the target domain, our approach shows an improvement upto $11.36 \%$, $6.65 \%$, $4.92\%$ for California, Floria and New York as target domain, respectively in terms of accuracy in predictions with respect to non-adaptive approach.
\end{itemize}

% CA - 11.36, 4.82
% NY - 4.92, 2.54
% FL - 6.65, 3.92

\section{Related Works}
In recent times, machine learning methods have increasingly found their way into solar power generation forecasting. A comparative analysis of traditional statistical and machine learning methods for solar PV generation forecasting was presented in \cite{b5}, where Convolutional Neural Network (CNN)-based hybrid models with occasional ensemble techniques were found to achieve superior performance in general. A hybrid model augmenting a Long Short-Term Memory (LSTM) model with Extreme Learning Machine (ELM) is proposed for distributed solar power generation forecasting \cite{b6}. Data from distributed photovoltaic (PV) plants are analyzed using the Spearman correlation coefficient to identify the most influential features affecting solar generation. The ELM component generalizes the underlying patterns, while the LSTM adaptively adjusts predictions across timesteps. The model's performance is evaluated under four distinct weather conditions: sunny, cloudy, rainy, and highly variable weather. The results obtained are compared with a backpropagation neural network (BPNN), an Elman neural network (ENN), and standalone LSTM and ELM models. Results demonstrate that the proposed LSTM-ELM hybrid achieves superior prediction accuracy across all test cases. However, the distributed PV plant data is sourced from rooftop systems in a single location of Southern China. The aggregated power output of spatially dispersed PV plants is predicted using a hybrid CNN model by introducing discrete wavelet transformation to a standard CNN model \cite{b7}. This approach achieves a 15\% improvement in accuracy over a regular CNN and a 7\% improvement over other machine learning and deep learning models when combined with wavelet transformation. Although, the PV plants are dispersed spatially, all of the sites are located within New South Wales region in Australia.  
%In \cite{b8}, a more robust short-term (24-hour) forecasting method is proposed. The authors employ the k-medoids method to identify four distinct intraday fluctuations and utilize similarity points among them for numerical weather prediction. They further align the joint distribution of selected weather variables using transfer component analysis (TCA) and apply support vector regression (SVR) for prediction. This method demonstrates superior performance compared to other approaches such as SVM, GBR, ANN, and ELM. However, while domain adaptation is incorporated, the analysis is limited to a single PV station within one climatic region. The authors in \cite{b9} discuss a type of Transfer Learning (TL) technique that trains a model with a source dataset and fine-tunes the parameters using the target dataset. Here they apply attention mechanism to assign weights to different input variables for each time step, while Dilated CNN and BiLSTM sub-components extract the spatial and temporal features of the data, respectively. Though the TL strategy is applied to two different sites of a PV station, they both are located within the same climatic region as they are from the same PV station. 

The work in \cite{b10} introduces a cluster-based Multi-Source Domain Adaptation (MSDA) framework to improve wind power forecasting for new wind farms. By transferring knowledge from operational farms, the approach reduces regression error by 20.63\% over instance-based MSDA baselines. In \cite{b11}, the author presents an online domain adaptive learning method for solar power forecasting that adapts its model structure in response to fluctuating weather conditions. This approach effectively monitors changes in data distribution and delivers reliable predictions without the need for labeled test data. Moreover, numerous studies in recent years have investigated domain adaptation models, particularly about improving their effectiveness for renewable energy sources. These efforts seek to not only boost the performance of these models but also to enhance their ability to tackle the specific challenges associated with renewable energy domains \cite{b8,b9,b12,b13,b14,b15,b16}. In \cite{b13} In this paper, we address the challenges of transferring knowledge between different domains in machine learning, particularly due to system discrepancies and limited training data, by implementing a knowledge transfer system (KTS) for long-term voltage stability assessment in power grids. They tried to convert system behaviors into heatmaps using a generalized method to prevent negative transfer and then apply a deep domain adaptation network (DDAN) that learns domain-invariant representations with strong semantic separation by incorporating a maximum mean discrepancy calculator. In \cite{b14}, authors present an adaptive transfer learning framework based on extreme gradient boosting for photovoltaic power forecasting, crucial in renewable-heavy power systems. The framework trains multiple models on source domain data and analyzes transferability by comparing SHAP value distributions. The most similar model is selected for knowledge transfer to the target domain.  Authors in \cite{b15} examine multi-step hourly solar irradiance forecasting using Multi-Model (MM) and Multi-Output (MO) approaches with feed-forward (FF) and LSTM neural networks.  They explore the domain adaptation on data from various regions, showing comparable performance for both FF and LSTM models, suggesting the feasibility of a single regional predictor. In \cite{b16}, the author introduces an asymptotic domain adaptive detection method that uses image style transfer to create an approximate domain and employs a multi-layer domain transfer method with a spatial attention mechanism to focus on abnormal features. 

However, these prior studies focus exclusively on individual geographic locations, rather than exploring how domain adaptation can address the variability in weather features caused by spatial differences. In \cite{b28}, the authors present a deep learning method to predict solar power generation based on weather data where the model is based on training a model in a fully supervised manner using data from a "source domain" (one country) and then adapting this model to a "target domain" (another country) where no data is available. The model, trained as a classification problem using a deep neural network (DNN) with convolutional layers and ReLU activations, classifies wind power into bins. Only the last layers of the pre-trained model are updated for the target domain, reducing computational cost and speeding up training. Similarly, fully supervised domain adaptation methods have been used to predict wind power \cite{b27}, building load \cite{buidling_load} and rain precipitation \cite{b29}. These works tried to address this research gap through the use of a source-free domain-adaptive deep learning framework where the solar PV production can be estimated for a region by deploying the model that has been developed by making use of data from another region. Both the regions are significantly different in terms of their seasonal behaviors and geographically far apart.
\section{Methodology}

% \begin{figure}
% \centering
% \includegraphics[width=\columnwidth]{images/hist_old.png}
% \caption{Histogram on solar power generation shown in five bins. The first bin (low power) has highest number of frequency. Because out of 24 hrs of a day around 12 hrs (6 pm to 6 am), the sunlight is not present. Hence no solar power is generated in that period which makes the highest number of frequency in the first bin.}
% \label{hist}
% \end{figure}

\begin{figure*}
\centering
\includegraphics[width=\textwidth]{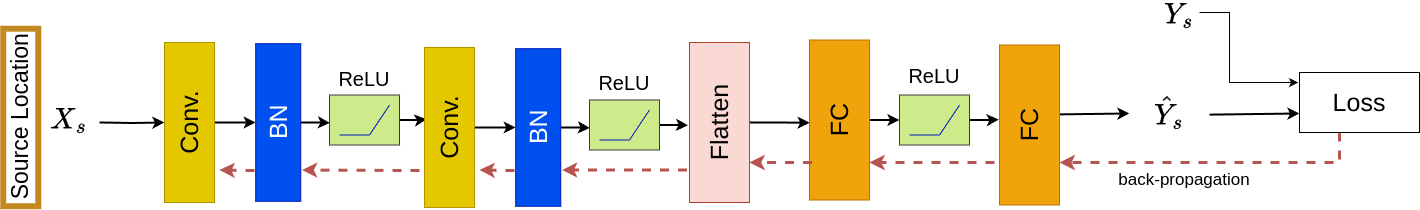}
\caption{Training on the source location: a model is trained from the scratch on the source side using the source data $(X_s, Y_s)$. Ather completion of the training the pretrained model is transferred to the target side. The same diagram has been used in \cite{b28}}
\label{source_training}
\end{figure*}

\begin{figure*}
\centering
\includegraphics[width=0.7\textwidth]{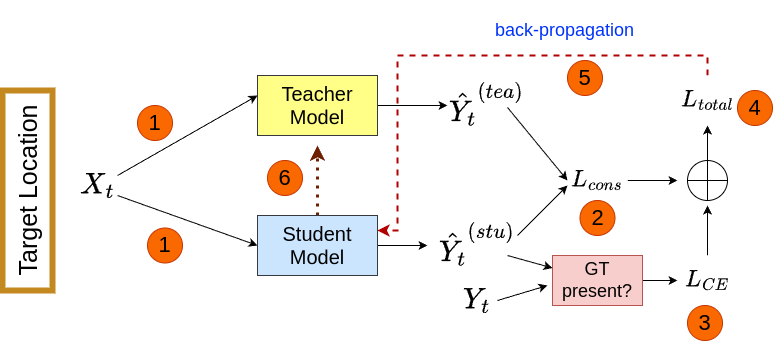}
\caption{ Source-Free Semi-supervised adaptation on the target side: The pre-trained weights of source model ($\theta_s$) is initialized for both teacher and student model. (1) Target data $X_t$ is fed into both models. (2) Using the prediction of both networks unsupervised loss $L_{cons}$ is calculated. (3) Supervised loss $L_{CE}$ is calculated for the samples which have ground truth in the target domain. (4) and (5) $L_{total}$ is used to update the student model by back-propagation. (6) The teacher model is updated by EMA method.
% \textcolor{red}{theta_s in the figure}
}
\label{source_free_adapt}
\end{figure*}

This paper proposes a method to predict the PV solar power generation of a particular location based on its meteorological characteristics using a pre-trained model of a different location. Therefore, this method utilizes two sets of datasets that are referred to as the source domain and the target domain, respectively. The source domain involves the domain where the model is trained in a fully supervised manner. When the training on the source domain is complete, only then the pre-trained model is deployed to the target domain. We assume that unlike the source side where we have annotation of the entire dataset, the target side has only a portion of annotated data. During adaptation, we shall consider source-free condition which refers to the unavailability of data and annotation from source side during adaptation. Therefore, the two key steps of our work are training a model on the source domain and adaptation of a pre-trained model on the target domain under source-free condition. 
 
 % (b) with source condition: In this setting, we consider that the annotated dataset of source domain is present during adaptation.
 
 % Fig. \ref{source_training}, \ref{source_free_dgm} and \ref{with_source_dgm} shows the total workflow of our approach.

% In order to make the process storage efficient, we assume that data of the source domain is not available on the target side. On the target domain, the pre-trained source model is adapted just by updating a few layers of the pre-trained model which makes the overall procedure faster and reduces computational cost.

% \begin{figure*}
% \centering
% \includegraphics[width=0.9\textwidth]{images/with_source_dgm.png}
% \caption{With Source Semi-supervised adaptation on the target side: The pre-trained source model is initialized for both teacher and student model. (1) Target data $X_t$  and Source data $X_s$ is fed into both models. (2) Using the prediction of both networks unsupervised loss $L_{cons}$ is calculated. (3) Supervised loss $L_{CE}$ is calculated for all the source domain samples and the target domain samples which have ground truth (4) and (5) $L_{total}$ is used to update the student model by back-propagation. (6) The teacher model is updated by EMA method. }
% \label{with_source_adapt}
% \end{figure*}

\subsection{Problem Formulation}

% \subsection{With-Source Domain Adaptation}
Given a source domain $D_s = \{(x_s^i, y_s^i)\}_{i=1}^{n_s}$, where $x_s \in \mathcal{X}_s$ are source domain samples and $y_s$ are the corresponding labels, and a target domain $D_t = \{(x_t^i, y_t^i)\}_{i=1}^{n_{ta}} \cup \{x_t^j\}_{j=1}^{n_{tu}}$, where $x_t \in \mathcal{X}_t$ are target domain samples and a subset of samples $n_{ta}$ have annotations ($\{(x_t^i, y_t^i)\}$ ), rest of the samples are not annotated $(x_t^j)$. We denote annotated target domain data as $D_t^a = \{(x_t^i, y_t^i)\}_{i=1}^{n_{ta}}$ , unannotated target domain data as $ D_t^u = \{x_t^j\}_{j=1}^{n_{tu}} $ where $ n_t = n_{t_a} + n_{t_u} $ and $D_t = D_t^a \cup D_t^u$. The source data is trained on a deep learning network and the pretrained model  weights can be denoted by $\theta_{s}$. The goal is to adapt the model $\theta_{s}$ to the target domain $D_t$. The challenge is that $D_s$ and $D_t$ follow different distributions, i.e., $P(\mathcal{X}_s) \neq P(\mathcal{X}_t)$, and the objective is to minimize the target prediction error by leveraging only $D_t$ during training.

\subsection{Training Model on Source Domain}

%\textcolor{red}{Paraphrase:} 

Source data have been utilized to train a feed-forward deep convolutional neural network. This prediction task has been formulated as a classification problem. The range of PV solar power generation was divided into $C$ number of bins (here $C = 5$) as executed in \cite{b28}. The source data is split into training ($70 \%$), validation ($15 \%$)  and testing data ($15 \%$). The deep neural network used for this work consists of convolutional layers (Conv.), batch normalization layers (BN) and fully connected layers (FC) as shown in Fig. \ref{source_training}. The final output layer is a FC layer having $C$ number of nodes to perform the mentioned prediction task. Input $X_{s}$ denotes the weather features of source domain and the output of the network is denoted by $\hat{Y}_{s}$ which is governed by the cross-entropy (CE) loss \cite{zhang2018generalized} using the ground truth of source side $Y_{s}$.\\

% \subsection{Training the Source Model}
% We shall train a deep neural network using the source data. We know that neural networks can be trained for two types of problems which are classification and regression. We formulate our problem as a regression problem as we want to predict the value of precipitation.  A feed-forward deep neural network of Fig. \ref{model} comprising 2 stages of hidden layers is used for training. The output layer has one node in order to facilitate regression. If the output of the network (predicted precipitation) is $Y_{src}$ and the ground truth value is $\hat{Y}_{src}$, the mean square error (MSE) loss function can be expressed by 

% \begin{equation}
% L_{src} =    \frac{1}{N}\sum_{i=1}^{N} (Y_{src} - \hat{Y}_{src})^{2}
% \end{equation}

% where $N$ indicates the number of samples in training.

\subsection{Adaptation on Target Domain} Adaptation on the target side is a training process where few samples labelled from target domain is used to adapt the model on the target domain. In the source-free approach (Fig. \ref{source_free_adapt}), the labelled data from the source side is not needed. We employ a teacher-student model \cite{wang2021knowledge} in both of the adaptation schemes. In a teacher-student model, there are two models where one acts as a teacher and another acts as a student model. The student model is guided by an unsupervised consistency loss ($L_{cons}$) and a supervised loss ($L_{CE}$) and updated via back-propagation \cite{lecun1988theoretical}. On the other hand, the teacher model is updated by Exponential Moving Average (EMA) \cite{hansun2013new} method.

We consider that source data and labels are unavailable. Only the pre-trained model weights $\theta_S$ from the source domain is available. Fig. \ref{source_free_adapt} illustrates the source-free setting. The teacher ($f_{tea}$) and student ($f_{stu}$) both model are initialized by the pre-trained model of source domain $\theta_s$. Both of the models are fed by same input of target domain $X_t$. There are two losses associated with the teacher student model. One is unsupervised consistency loss ($L_{cons}$)which is calculated by the loss between the prediction of teacher model and student model. And another is supervised crossentropy loss ($L_{CE}$ which are obtained only for the samples where target domain is labelled. Hence $L_{cons}$ can be obtained for all samples of target domain. However, $L_{CE}$ can be obtained only for a few samples.

\begin{equation}
    L_{cons} = -\sum_{i \in D_t} \sum_{c=1}^{C} \hat{Y}_{t_i}^{(tea)c} \log(\hat{Y}_{t_i}^{(stu)c}) 
    \label{lcons}
\end{equation}

\begin{equation}
L_{CE} = -\sum_{i \in D_t^a}  \sum_{c=1}^{C} Y_{t_i}^c \cdot \log \hat{Y}_{t_i}^{(stu)c}
\label{loss_ce}
\end{equation}

The student model weight ($\theta_{stu}$) is updated by the total loss through back-propagation. 

\begin{equation}
L_{total} = L_{cons} + \lambda L_{CE}
\label{loss_total}
\end{equation}

The teacher model is updated by the EMA update using the student model.

\begin{equation}
\theta_{tea} \leftarrow \alpha \cdot \theta_{tea} + (1 - \alpha) \cdot \theta_{stu}
\label{ema}
\end{equation}

where $\alpha$ is the EMA decay coefficient and $0 \leq \alpha < 1$. Algorithm \ref{alg:teacher_student_da} demonstrates the overall approach

% \subsubsection{With-Source Adaptation}

% \begin{equation}
% \footnotesize
%     L_{cons} = - \sum_{i=1}^{N} \sum_{c=1}^{C} 
%     \hat{Y}_{t_i}^{(tea)c} \log(\hat{Y}_{t_i}^{(stu)c})
%     - \sum_{i=1}^{N} \sum_{c=1}^{C} 
%     \hat{Y}_{s_i}^{(tea)c} \log(\hat{Y}_{s_i}^{(stu)c})
% \end{equation}

% \begin{equation}
% \footnotesize
%     L_{CE} = -\sum_{i \in D_t^a}  \sum_{c=1}^{C} 
%     Y_{t_i}^c \cdot \log \hat{Y}_{t_i}^{(stu)c} 
%     + \sum_{i \in D_s}  \sum_{c=1}^{C} 
%     Y_{s_i}^c \cdot \log \hat{Y}_{s_i}^{(stu)c}
% \end{equation}

\begin{algorithm}[t]
\caption{Semi-Supervised Domain Adaptation using Teacher-Student Model}
\label{alg:teacher_student_da}
    \begin{algorithmic}[1]

    \Require $D_s$ (Source Domain Data), $D_t^a$ (Annotated Target Data), $D_t^u$ (Unannotated Target Data), $f_{tea}$ (Teacher Model), $f_{stu}$ (Student Model), $L_{cons}$ (Consistency Loss), $L_{CE}$ (Cross-Entropy Loss), $\alpha$ (EMA Decay Coefficient)
    \Ensure $\theta_s \leftarrow$ Trained weights on $D_s$
    
    \State \textbf{Initialize:} Teacher model $f_{tea}$ and Student model $f_{stu}$ with pre-trained weights $\theta_s$
    
    \While{not converged}
        \State Sample a batch $X_t$ comprising elements from $\{(x_t^a, y_t^a)\} \in D_t^a$ and $\{x_t^u\} \in D_t^u$
        
        \State \textbf{Step 1: Feed $X_t$ to both Teacher and Student models}
        \State Compute teacher predictions on $X_t$:  $\hat{Y}^{(tea)}_t = f_{tea}(X_t)$
        \State Compute student predictions on $X_t$:  $\hat{Y}^{(stu)}_t = f_{stu}(X_t)$
    
        \State \textbf{Step 2: Loss Computation}
        \State Compute consistency loss $L_{cons}$ (eq. \ref{lcons}) between the predictions $\hat{Y}^{(tea)}_t$ and $\hat{Y}^{(stu)}_t$ for all samples in $X_t$
        \State Compute cross-entropy loss $L_{CE}$ (eq. \ref{loss_ce}) for the samples having annotations $\{(x_t^a, y_t^a)\} \in D_t^a$
        
        \State \textbf{Step 3: Update Student Model}
        \State Update the Student model $f_{stu}$ using back-propagation based on the total loss (eq. \ref{loss_total})
    
        \State \textbf{Step 4: Update Teacher Model}
        \State Update the Teacher model $f_{tea}$ using Exponential Moving Average (EMA) update (eq. \ref{ema}).
    
    \EndWhile
    
    \Return Teacher model $f_{tea}$
    
    \end{algorithmic}
\end{algorithm}

\begin{table}[b]
    \caption{Comparison of accuracies between different methods in predicting solar power generation using source data for California (CA), Florida (FL) and New York (NY).}
    \begin{center}
    % c|c|
    \begin{tabular}{|c|c|c|c|}
    \hline \textbf{Classifier Method} & \textbf{CA} & \textbf{FL} & \textbf{NY} \\
    \hline Adaboost & $76.65 \%$ & $68.58 \%$ & $67.75 \%$ \\
    \hline Gradient Boosting & $78.99 \%$ & $73.14 \%$ & $70.66 \%$ \\
    \hline Random Forest& $76.71 \%$ & $68.78 \%$ & $67.92 \%$ \\
    \hline Deep Neural Network& \textbf{$81.02 \%$} & \textbf{$80.99 \%$} & \textbf{$80.59 \%$} \\
    \hline

        \end{tabular}
    
    \label{table:source}
    \end{center}
\end{table}

\section{Experiment and Results}

\begin{table*}[!ht]
  \caption{Comparing the performance in terms of Accuracy for Deep Learning Network on different locations.}
  \small\centering
    \begin{tabular}{|c|c|c| c c c c c|}
    \hline
    \multirow{2}{*}{\textbf{Source}} & \multirow{2}{*}{\textbf{Target}} & \multirow{2}{*}{\textbf{w/o adapt.}} 
    & \multicolumn{5}{c|}{\textbf{Annotation amount in target domain $(p \%)$}} \\
    \cline{4-8}
    & & & \textbf{0\%} & \textbf{10\%} & \textbf{20\%} & \textbf{50\%} & \textbf{100\%} \\
    \hline
    \textbf{CA} 
    & NY & 72.68\% & 73.10\% & 77.11\% & 77.60\% & 78.30\% & 79.94\% \\
    & FL & 69.43\% & 69.91\% & 74.03\% & 76.08\% & 77.42\% & 81.56\% \\
    
    \hline
    \textbf{FL} 
    & CA & 64.09\% & 68.80\% & 74.05\% & 75.45\% & 77.62\% & 82.99\% \\
    & NY & 74.88\% & 75.31\% & 76.82\% & 77.42\% & 77.62\% & 79.99\% \\
    \hline
    \textbf{NY} 
    & CA & 72.77\% & 75.68\% & 77.48\% & 77.59\% & 78.01\% & 83.24\% \\
    & FL & 74.06\% & 74.91\% & 77.28\% & 77.98\% & 78.85\% & 81.50\% \\
    \hline
    \end{tabular}
\label{table:adapt_restructured}
\end{table*}

\subsection{Dataset}

Solar generation is predicted through utilization of different weather features. The solar data is retrieved from \cite{b17} and the meteorological data is collected from \cite{b18}. The solar dataset provides synthetic PV generation data for different states that serves as a realistic representation of specific geographic coordinates. The dataset contains solar generation data for a plant of a given kW capacity in 5-minute intervals for the year 2006. Solar production data is retrieved for the three states of the USA: California (CA), Florida (FL), and New York (NY). Validation across these geographically dispersed datasets, which represent substantially different atmospheric regimes, will demonstrate the robustness and practical value of our approach. The weather dataset includes the following nine features: direct normal irradiance (DNI), direct horizontal irradiance (DHI), global horizontal irradiance (GHI), dew point, temperature, pressure, relative humidity, wind direction, wind speed, and surface albedo. However, the weather data available for the above-mentioned locations and time period is at 30-minute intervals. Therefore, to keep both datasets consistent, solar production data is averaged to 30-minute interval data to match the temporal resolution of meteorological data. This process yields 17,520 samples, representing one year's worth of solar generation data at 30-minute intervals for each location. Here, weather data act as the primary features for predicting solar output through the domain adaptation algorithm.

The random forest algorithm \cite{b19} is widely used for any classification or regression task to identify the key parameters and help in eliminating irrelevant features. In this work, we also apply this algorithm to find the impactful features for predicting solar power, and only a reduced number of applicable features (DNI, DHI, GHI, temperature, wind direction, and wind speed) are utilized to accomplish the prediction task.

\subsection{Results}
In this sub-section, different types of analyses are presented on our proposed method. Firstly, the performance of the deep learning model trained on the source data will be compared against other ensembling machine learning approaches. Secondly, the quantitative performance of the models for adaptation and non-adaptation scenarios will be evaluated as well. We also demonstrate the adaptation performance under different annotation amounts in the target domain. 

\subsubsection{Performance of Training Source Model from Scratch}
% \textcolor{red}{Paraphrase}\\
% We train the deep learning model using source data in order to get the pretrained model. Our implementation is executed by the popular deep learning framework Pytorch \cite{b26}. We use the learning rate $10^{-4}$, batch size $1000$, and popular optimization technique ADAM \cite{b25} optimizer. 

% We compare the performance of deep learning model with some standard ensembling machine learning techniques such as Adaboost Classifier \cite{b21} , Gradient Boosting Classifier \cite{b22}, Random Forest Classifier \cite{b23}. TABLE \ref{table:source} illustrates that Deep Neural Network Classifier outperforms all other ensembling machine learning methods. Deep learning method improves the result at least $2.03 \%$, $ 7.85 \%$ and $9.93 \%$ for states California (CA), Florida (FL), and New York (NY), respectively. Hence choosing a deep neural network in our task was the most reasonable choice. If the hyper-parameters of the deep learning model are fine-tuned, we may achieve slightly better accuracy. 

Initially, the deep learning model is trained using the source data to achieve the pretrained model. The training is implemented through utilization of widely used deep learning framework, PyTorch \cite{b26}. A learning rate of $10^{-4}$ and a batch size $1000$ is adopted. One of the most prominent optimization techniques, ADAM optimizer has been applied in this work \cite{b25}. 

To validate the effectiveness of the deep learning model, we compare its performance with standard ensemble machine learning techniques such as the Adaboost Classifier \cite{b21}, Gradient Boosting Classifier \cite{b22}, and Random Forest Classifier \cite{b23}. It is evident from the results in Table \ref{table:source} that the Deep Neural Network Classifier surpasses all other ensemble machine learning approaches. By comparing the results, we observe that the deep learning approach increases the training accuracy by at least $2.03\%$, $7.85\%$, and $9.93\%$ for the states of California (CA), Florida (FL), and New York (NY), respectively. Thus, selecting a deep neural network for our task was the most appropriate choice. We could achieve even higher accuracy through the process of fine-tuning the hyperparameters of the deep learning model.

\subsubsection{Performance of Source-Free Adaptation}
% \textcolor{red}{Elaborate}
% In TABLE \ref{table:adapt_restructured}, we demonstrate the performance of adaptation. We show the performance of different locations without adaptation. Then we show the performance of adaptation based on different amounts ($p \%$) of annotation in the training data of target domain. $p=0 \text{ and} 100 $ mean fully-unsupervised and fully-supervised settings respectively. Rest of the values of p indicate the semi-supervised setting. We note that using only $p = 10$ data in the training, we get a  performance boost of $4.92 \%$ and $6.65 \%$ in adaptation from CA to NY and FL ,respectively. In addition, we note that compared to $p=10$ we obtain a performance boost of $2.34 \%$ and $5.48 \%$ for $p=100$. But $p=10$ means there is $90 \%$ less annotation cost compared to $p=100$. Therefore, it can be concluded that our proposed approach substantially reduces annotation costs with a far less trade-off in performance degradation.

TABLE \ref{table:adapt_restructured} illustrates the effectiveness of domain adaptation in predicting solar generation across different locations under varying levels of supervision. A total of six source-target location combinations are explored, with each of the three states acting as a source location while the other two serve as target locations. For each combination, prediction accuracy without adaptation is first presented. These results are then compared using domain adaptation with different levels of supervision. The parameter $p$ represents the percentage of annotated training data in the target dataset, indicating the supervision level in our domain adaptation. Here, $p=0\%$ denotes fully unsupervised domain adaptation, while $p=100\%$ represents fully supervised domain adaptation. Intermediate values of $p$ indicate semi-supervised learning with varying levels of supervision.

When comparing prediction accuracies without domain adaptation to those of fully unsupervised domain adaptation, we observe minimal improvement in most cases. This increase is typically less than 1\%, which is expected given the lack of annotated training data in fully unsupervised adaptation. At $p=10\%$, performance improves by 4.92\% and 6.65\% when adapting from source location CA to target locations NY and FL, respectively. As $p$ increases, prediction accuracy also improves, as additional annotated data allows the deep learning model to learn more effectively. Comparing the results for $p=10\%$ with $p=100\%$ (fully supervised domain adaptation), we note only a 2.34\% and 5.48\% performance increase for CA-NY and CA-FL combinations, respectively. With $p = 10$, annotation costs are reduced by 90\% compared to $p = 100$. Thus, while our proposed approach shows limited improvement in fully unsupervised adaptation, it substantially reduces annotation costs with minimal performance degradation as a trade-off.

\section{Conclusion}

% In this study, we presented a semi-supervised, source-free deep domain adaptation framework for accurate and location-agnostic solar power prediction across diverse climatic regions. By training a deep learning model on source domain data and adapting it to target domains with limited labeled data, our approach effectively addresses the domain shift challenges inherent to solar power forecasting. Using a teacher-student model, we achieved significant accuracy improvements, even in scenarios with minimal target domain annotations, thereby reducing annotation costs while maintaining high prediction performance. Experimental results on datasets from California, Florida, and New York validate the robustness of our model, showing up to 11.36 $\%$ accuracy gains over non-adaptive methods. Furthermore, our source-free adaptation setup enhances computational efficiency and ensures data privacy, making it well-suited for practical deployment across geographically diverse regions. This work underscores the potential of domain adaptation in renewable energy applications, setting the groundwork for further advancements in scalable, cost-effective solar power prediction.

% \textcolor{red}{Change a bit}
In this study, we present a semi-supervised, source-free deep domain adaptation framework for accurate, location-agnostic solar power prediction across geographically dispersed sites with varying meteorological conditions, requiring minimal target-domain supervision. Our approach effectively addresses the domain shift challenge observed in solar power forecasting across geographically distant locations by training a deep learning model on source domain data and adapting it to target domains with limited or no labeled data. The teacher-student model employed here achieves significant accuracy improvements even with only a small amount of annotated target data, providing high prediction performance with reduced annotation costs. Experimental results on datasets from California, Florida, and New York validate the robustness of our model, demonstrating up to 11.36\% accuracy gains over non-adaptive methods. Additionally, our source-free adaptation setup enhances computational efficiency and ensures data privacy, making it suitable for practical applications. This work highlights the application and benefits of domain adaptation in advancing renewable energy prediction.

% \section{Declaration of generative AI and AI-assisted technologies in the writing process}

% During the preparation of this work the authors used ChatGPT in order to ensure grammatical correctness and clarity of writing. After using this tool/service, the authors reviewed and edited the content as needed and take full responsibility for the content of the publication.
% 

% if have a single appendix:
%\appendix[Proof of the Zonklar Equations]
% or
%\appendix  % for no appendix heading
% do not use \section anymore after \appendix, only \section*
% is possibly needed

% use appendices with more than one appendix
% then use \section to start each appendix
% you must declare a \section before using any
% \subsection or using \label (\appendices by itself
% starts a section numbered zero.)
%

% \appendices
% \section{Proof of the First Zonklar Equation}
% Appendix one text goes here.

% % you can choose not to have a title for an appendix
% % if you want by leaving the argument blank
% \section{}
% Appendix two text goes here.

% % % use section* for acknowledgment
% \section*{Declaration of generative AI and AI-assisted technologies in the writing process}

% During the preparation of this work the authors used ChatGPT in order to ensure grammatical correctness and clarity of writing. After using this tool/service, the authors reviewed and edited the content as needed and take full responsibility for the content of the publication.

% The authors would like to thank...

% Can use something like this to put references on a page
% by themselves when using endfloat and the captionsoff option.
\ifCLASSOPTIONcaptionsoff
  \newpage
\fi

% trigger a \newpage just before the given reference
% number - used to balance the columns on the last page
% adjust value as needed - may need to be readjusted if
% the document is modified later
%\IEEEtriggeratref{8}
% The "triggered" command can be changed if desired:
%\IEEEtriggercmd{\enlargethispage{-5in}}

% references section

% can use a bibliography generated by BibTeX as a .bbl file
% BibTeX documentation can be easily obtained at:
% http://mirror.ctan.org/biblio/bibtex/contrib/doc/
% The IEEEtran BibTeX style support page is at:
% http://www.michaelshell.org/tex/ieeetran/bibtex/
%\bibliographystyle{IEEEtran}
% argument is your BibTeX string definitions and bibliography database(s)
%\bibliography{IEEEabrv,../bib/paper}
%
% <OR> manually copy in the resultant .bbl file
% set second argument of \begin to the number of references
% (used to reserve space for the reference number labels box)
% \begin{thebibliography}{1}

% \bibitem{IEEEhowto:kopka}
% H.~Kopka and P.~W. Daly, \emph{A Guide to \LaTeX}, 3rd~ed.\hskip 1em plus
%   0.5em minus 0.4em\relax Harlow, England: Addison-Wesley, 1999.

% \end{thebibliography}

\bibliographystyle{IEEEtran}  % or any other style you prefer
\bibliography{IEEEexample}     % 'references' is the name of your .bib file

% Generated by IEEEtran.bst, version: 1.14 (2015/08/26)
\begin{thebibliography}{10}
\providecommand{\url}[1]{#1}
\csname url@samestyle\endcsname
\providecommand{\newblock}{\relax}
\providecommand{\bibinfo}[2]{#2}
\providecommand{\BIBentrySTDinterwordspacing}{\spaceskip=0pt\relax}
\providecommand{\BIBentryALTinterwordstretchfactor}{4}
\providecommand{\BIBentryALTinterwordspacing}{\spaceskip=\fontdimen2\font plus
\BIBentryALTinterwordstretchfactor\fontdimen3\font minus \fontdimen4\font\relax}
\providecommand{\BIBforeignlanguage}[2]{{%
\expandafter\ifx\csname l@#1\endcsname\relax
\typeout{** WARNING: IEEEtran.bst: No hyphenation pattern has been}%
\typeout{** loaded for the language `#1'. Using the pattern for}%
\typeout{** the default language instead.}%
\else
\language=\csname l@#1\endcsname
\fi
#2}}
\providecommand{\BIBdecl}{\relax}
\BIBdecl

\bibitem{b1}
L.~Capuano, ``U.s. energy information administration's international energy outlook 2020,'' U.S. Department of Energy: Washington, D.C., Report No. 7, 2020.

\bibitem{b2}
Y.~Abdelilah, H.~Bahar, T.~Criswell, P.~Bojek, F.~Briens, and P.~L. Feuvre, ``Renewables 2020: Analysis and forecast to 2025,'' International Energy Agency (IEA): Paris, France, 2020.

\bibitem{irenaSolarEnergy}
``{S}olar energy --- irena.org,'' \url{https://www.irena.org/Energy-Transition/Technology/Solar-energy}, [Accessed 25-10-2024].

\bibitem{ieaSolar}
``{S}olar - {I}{E}{A} --- iea.org,'' \url{https://www.iea.org/energy-system/renewables/solar-pv }, [Accessed 25-10-2024].

\bibitem{stacke2020measuring}
K.~Stacke, G.~Eilertsen, J.~Unger, and C.~Lundstr{\"o}m, ``Measuring domain shift for deep learning in histopathology,'' \emph{IEEE journal of biomedical and health informatics}, vol.~25, no.~2, pp. 325--336, 2020.

\bibitem{b27}
M.~S. Islam~Sajol, M.~Shazid~Islam, A.~S.~M. Jahid~Hasan, M.~Saydur~Rahman, and J.~Yusuf, ``Wind power prediction across different locations using deep domain adaptive learning,'' in \emph{2024 6th Global Power, Energy and Communication Conference (GPECOM)}, 2024, pp. 518--523.

\bibitem{b28}
M.~S. Islam, A.~S. M.~J. Hasan, M.~S. Rahman, J.~Yusuf, M.~S.~I. Sajol, and F.~A. Tumpa, ``Location agnostic source-free domain adaptive learning to predict solar power generation,'' in \emph{2023 IEEE International Conference on Energy Technologies for Future Grids (ETFG)}, 2023, pp. 1--6.

\bibitem{b29}
M.~S. Islam, M.~S. Rahman, M.~S. Ul~Haque, F.~A. Tumpa, M.~S. Bin~Hossain, and A.~A. Arabi, ``Location agnostic adaptive rain precipitation prediction using deep learning,'' in \emph{2023 IEEE 9th International Women in Engineering (WIE) Conference on Electrical and Computer Engineering (WIECON-ECE)}, 2023, pp. 148--153.

\bibitem{Tipu}
\BIBentryALTinterwordspacing
M.~S. Islam, M.~A.~T. Rony, and T.~Sultan, ``Gastrovrg: Enhancing early screening in gastrointestinal health via advanced transfer features,'' \emph{Intelligent Systems with Applications}, vol.~23, p. 200399, 2024. [Online]. Available: \url{https://www.sciencedirect.com/science/article/pii/S2667305324000747}
\BIBentrySTDinterwordspacing

\bibitem{PINTO}
\BIBentryALTinterwordspacing
G.~Pinto, Z.~Wang, A.~Roy, T.~Hong, and A.~Capozzoli, ``Transfer learning for smart buildings: A critical review of algorithms, applications, and future perspectives,'' \emph{Advances in Applied Energy}, vol.~5, p. 100084, 2022. [Online]. Available: \url{https://www.sciencedirect.com/science/article/pii/S2666792422000026}
\BIBentrySTDinterwordspacing

\bibitem{sumon}
\BIBentryALTinterwordspacing
R.~I. Sumon, H.~Ali, S.~Akter, S.~M.~I. Uddin, M.~A.~I. Mozumder, and H.-C. Kim, ``A deep learning-based approach for precise emotion recognition in domestic animals using efficientnetb5 architecture,'' \emph{Eng}, vol.~6, no.~1, 2025. [Online]. Available: \url{https://www.mdpi.com/2673-4117/6/1/9}
\BIBentrySTDinterwordspacing

\bibitem{hoffman2018algorithms}
J.~Hoffman, M.~Mohri, and N.~Zhang, ``Algorithms and theory for multiple-source adaptation,'' \emph{Advances in neural information processing systems}, vol.~31, 2018.

\bibitem{sambasivan2021everyone}
N.~Sambasivan, S.~Kapania, H.~Highfill, D.~Akrong, P.~Paritosh, and L.~M. Aroyo, ``“everyone wants to do the model work, not the data work”: Data cascades in high-stakes ai,'' in \emph{proceedings of the 2021 CHI Conference on Human Factors in Computing Systems}, 2021, pp. 1--15.

\bibitem{liu2022}
\BIBentryALTinterwordspacing
X.~Liu, C.~Yoo, F.~Xing, H.~Oh, G.~E. Fakhri, J.-W. Kang, and J.~Woo, ``Deep unsupervised domain adaptation: A review of recent advances and perspectives,'' 2022. [Online]. Available: \url{https://arxiv.org/abs/2208.07422}
\BIBentrySTDinterwordspacing

\bibitem{yu2023}
\BIBentryALTinterwordspacing
Y.-C. Yu and H.-T. Lin, ``Semi-supervised domain adaptation with source label adaptation,'' 2023. [Online]. Available: \url{https://arxiv.org/abs/2302.02335}
\BIBentrySTDinterwordspacing

\bibitem{b5}
\BIBentryALTinterwordspacing
R.~Ahmed, V.~Sreeram, Y.~Mishra, and M.~Arif, ``A review and evaluation of the state-of-the-art in pv solar power forecasting: Techniques and optimization,'' \emph{Renewable and Sustainable Energy Reviews}, vol. 124, p. 109792, 2020. [Online]. Available: \url{https://www.sciencedirect.com/science/article/pii/S1364032120300885}
\BIBentrySTDinterwordspacing

\bibitem{b6}
\BIBentryALTinterwordspacing
D.~Yongsheng, J.~Fengshun, Z.~Jie, and L.~Zhikeng, ``A short-term power output forecasting model based on correlation analysis and elm-lstm for distributed pv system,'' \emph{Journal of Electrical and Computer Engineering}, vol. 2020, p. 1–10, Jun. 2020. [Online]. Available: \url{http://dx.doi.org/10.1155/2020/2051232}
\BIBentrySTDinterwordspacing

\bibitem{b7}
S.~Almaghrabi, M.~Rana, M.~Hamilton, and M.~S. Rahaman, ``Spatially aggregated photovoltaic power prediction using wavelet and convolutional neural networks,'' in \emph{2021 International Joint Conference on Neural Networks (IJCNN)}, 2021, pp. 1--8.

\bibitem{b10}
S.~Tasnim, A.~Rahman, A.~Oo, and M.~Haque, ``Wind power prediction in new stations based on knowledge of existing stations: A cluster based multi source domain adaptation approach,'' \emph{Knowledge-Based Systems}, vol. 145, 12 2017.

\bibitem{b11}
H.~Sheng, B.~Ray, K.~Chen, and Y.~Cheng, ``Solar power forecasting based on domain adaptive learning,'' \emph{IEEE Access}, vol.~8, pp. 198\,580--198\,590, 2020.

\bibitem{b8}
J.~Wang, G.~Yan, M.~Ren, X.~Xu, Z.~Ye, and Z.~Zhu, ``Short term photovoltaic power prediction based on transfer learning and considering sequence uncertainty,'' \emph{Journal of Renewable and Sustainable Energy}, vol.~15, no.~1, 2023.

\bibitem{b9}
Y.~Tang, K.~Yang, S.~Zhang, and Z.~Zhang, ``Photovoltaic power forecasting: A hybrid deep learning model incorporating transfer learning strategy,'' \emph{Renewable and Sustainable Energy Reviews}, vol. 162, p. 112473, 2022.

\bibitem{b12}
X.~Wang, Q.~Kang, M.~Zhou, S.~Yao, and A.~Abusorrah, ``Domain adaptation multitask optimization,'' \emph{IEEE Transactions on Cybernetics}, vol.~53, no.~7, pp. 4567--4578, 2023.

\bibitem{b13}
H.~Cai and D.~J. Hill, ``Knowledge transfer for long-term voltage stability assessment between power grids based on deep domain adaptation networks,'' in \emph{2020 12th IEEE PES Asia-Pacific Power and Energy Engineering Conference (APPEEC)}, 2020, pp. 1--5.

\bibitem{b14}
Z.~Tang, Y.~Tang, A.~Qiao, J.~Liu, and J.~Gao, ``Transfer learning based photovoltaic power forecasting with xgboost,'' in \emph{2023 Panda Forum on Power and Energy (PandaFPE)}, 2023, pp. 1781--1785.

\bibitem{b15}
\BIBentryALTinterwordspacing
G.~Guariso, G.~Nunnari, and M.~Sangiorgio, ``Multi-step solar irradiance forecasting and domain adaptation of deep neural networks,'' \emph{Energies}, vol.~13, no.~15, 2020. [Online]. Available: \url{https://www.mdpi.com/1996-1073/13/15/3987}
\BIBentrySTDinterwordspacing

\bibitem{b16}
J.~Zhang, G.~Peng, R.~Song, S.~Zhang, Y.~Tan, T.~Pu, and J.~Wang, ``Asymptotic domain adaptive detection for abnormal targets in transmission lines under complex weather conditions,'' \emph{CSEE Journal of Power and Energy Systems}, pp. 1--13, 2023.

\bibitem{buidling_load}
A.~S. M.~J. Hasan, M.~S. Islam, M.~S. Rahman, M.~S.~I. Sajol, and J.~Yusuf, ``Deep learning based cross-location building load prediction using domain adaptation,'' in \emph{2024 IEEE International Conference And Exposition On Electric And Power Engineering (EPEi)}, 2024, pp. 362--367.

\bibitem{zhang2018generalized}
Z.~Zhang and M.~Sabuncu, ``Generalized cross entropy loss for training deep neural networks with noisy labels,'' \emph{Advances in neural information processing systems}, vol.~31, 2018.

\bibitem{wang2021knowledge}
L.~Wang and K.-J. Yoon, ``Knowledge distillation and student-teacher learning for visual intelligence: A review and new outlooks,'' \emph{IEEE transactions on pattern analysis and machine intelligence}, vol.~44, no.~6, pp. 3048--3068, 2021.

\bibitem{lecun1988theoretical}
Y.~LeCun, D.~Touresky, G.~Hinton, and T.~Sejnowski, ``A theoretical framework for back-propagation,'' in \emph{Proceedings of the 1988 connectionist models summer school}, vol.~1, 1988, pp. 21--28.

\bibitem{hansun2013new}
S.~Hansun, ``A new approach of moving average method in time series analysis,'' in \emph{2013 conference on new media studies (CoNMedia)}.\hskip 1em plus 0.5em minus 0.4em\relax IEEE, 2013, pp. 1--4.

\bibitem{b17}
N.~R. E.~L. (NREL), ``Solar power data for integration studies,'' \url{https://www.nrel.gov/grid/solar-power-data.html}, Accessed: October 16, 2024, online.

\bibitem{b18}
------, ``Weather data,'' \url{https://sam.nrel.gov/weatherdata.html}, Accessed: October 16, 2024, online.

\bibitem{b19}
\BIBentryALTinterwordspacing
M.~Hummon, E.~Ibanez, G.~Brinkman, and D.~Lew, ``Sub-hour solar data for power system modeling from static spatial variability analysis: Preprint,'' 12 2012. [Online]. Available: \url{https://www.osti.gov/biblio/1059579}
\BIBentrySTDinterwordspacing

\bibitem{b26}
S.~Imambi, K.~B. Prakash, and G.~Kanagachidambaresan, ``Pytorch,'' \emph{Programming with TensorFlow: solution for edge computing applications}, pp. 87--104, 2021.

\bibitem{b25}
D.~P. Kingma, ``Adam: A method for stochastic optimization,'' \emph{arXiv preprint arXiv:1412.6980}, 2014.

\bibitem{b21}
R.~E. Schapire, ``Explaining adaboost,'' in \emph{Empirical inference: festschrift in honor of vladimir N. Vapnik}.\hskip 1em plus 0.5em minus 0.4em\relax Springer, 2013, pp. 37--52.

\bibitem{b22}
P.~Prettenhofer and G.~Louppe, ``Gradient boosted regression trees in scikit-learn,'' in \emph{PyData 2014}, 2014.

\bibitem{b23}
T.~F. Cootes, M.~C. Ionita, C.~Lindner, and P.~Sauer, ``Robust and accurate shape model fitting using random forest regression voting,'' in \emph{Computer Vision--ECCV 2012: 12th European Conference on Computer Vision, Florence, Italy, October 7-13, 2012, Proceedings, Part VII 12}.\hskip 1em plus 0.5em minus 0.4em\relax Springer, 2012, pp. 278--291.

\end{thebibliography}

\end{document}